# Parameters Optimization of Deep Learning Models using Particle Swarm Optimization

Basheer Qolomany[1], Majdi Maabreh[1], Ala Al-Fuqaha[1], Ajay Gupta[1]
[1]Computer Science Dept., College of Engineering and Applied Sciences,
Western Michigan University,
Kalamazoo, Michigan, USA
{basheer.qolomany, majdi.a.maabreh, ala.al-fuqaha, ajay.gupta}@wmich.edu

Driss Benhaddou[2]
[2]Engineering Technology Dept., College of Technology, University of Houston, Houston, Texas, USA
dbenhadd@central.uh.edu

*Abstract*— **Deep learning has been successfully applied in several fields such as machine translation, manufacturing, and pattern recognition. However, successful application of deep learning depends upon appropriately setting its parameters to achieve high-quality results. The number of hidden layers and the number of neurons in each layer of a deep machine learning network are two key parameters, which have main influence on the performance of the algorithm. Manual parameter setting and grid search approaches somewhat ease the users' tasks in setting these important parameters. Nonetheless, these two techniques can be very time-consuming. In this paper, we show that the Particle swarm optimization (PSO) technique holds great potential to optimize parameter settings and thus saves valuable computational resources during the tuning process of deep learning models.**

**Specifically, we use a dataset collected from a Wi-Fi campus network to train deep learning models to predict the number of occupants and their locations. Our preliminary experiments indicate that PSO provides an efficient approach for tuning the optimal number of hidden layers and the number of neurons in each layer of the deep learning algorithm when compared to the grid search method. Our experiments illustrate that the exploration process of the landscape of configurations to find the optimal parameters is decreased by 77 % - 85%. In fact, the PSO yields even better accuracy results.**

## I. INTRODUCTION

Deep learning is an aspect of artificial neural networks that aims to imitate complex learning methods that human beings use to gain certain types of knowledge. We can think of deep learning as a technique that employs neural networks that utilize multiple hidden layers of abstraction between the input and output layers. This is in contrast to traditional shallow neural networks that employ one hidden layer [1].

Deep learning models are utilized in a wide variety of applications including the popular iOS Siri and Google voice systems. Recently, deep neural networks have been utilized to win numerous contests in pattern recognition and machine learning. Some leading examples include Microsoft research on a deep learning system that demonstrated the ability to classify 22,000 categories of pictures at 29.8 percent of accuracy. They also demonstrated real-time speech translation between Mandarin Chinese and English. [2]. Deep learning is made available by open source projects as well, currently various commonly used deep learning platforms include: $H_2O$ platform, Deeplearning4j (DL4j), Theano, Torch, TensorFlow, and Caffe.

One of the challenges in a successful implementation of deep machine learning is setting the values for its many parameters, particularly the topology of its network. Let L be the number of hidden layers, $N_i$ be the number of neurons in layer *i* and $N=\{N_1, N_2, …, N_L\}$. Parameters L and N are very important and have a major influence on the performance of deep machine learning. Manually tuning these parameters (essentially through trial and error method) and finding high-quality settings is a time-consuming process [3]. Besides, the solutions obtained by the manual process are usually not equally distributed in the objective space.

To address this challenge, grid search is a common approach for setting parameter values of the deep learning models. Grid search is more efficient and saves time in setting L and N; with this approach, a list of discrete values of L and N are prepared in advance, where each entry shows the number of hidden layers and its corresponding number of neurons. The deep learning algorithm trains multiple different models using all the list's entries. Finally, the selection of the parameters is measured using the models' accuracy. However, grid search is still a computationally demanding process as the number of possible combinations is exponential, especially when the number of parameters increases and the interval between discrete values is reduced. In addition if the list of parameters are poorly chosen, the network may learn slowly, or perhaps not at all [4].

This paper proposes another parameter selection method for deep learning models using PSO. PSO is a popular population-based heuristic algorithm that simulates the social behavior of individuals such as birds flocking, a school of fish swimming or a colony of ants moving to a potential position to achieve particular objectives in a multidimensional space [5]. PSO is found to have the extensive capability of global optimization for its simple concept, easy implementation, scalability, robustness, and fast convergence. It employs only simple mathematical operators and is computationally inexpensive in terms of both memory requirements and speed [6].

Several researchers have explored parameter optimization of various machine learning algorithms. PSO has been applied to train shallow neural networks [7]. There are a number of studies about specifying and optimizing the initial weights of Artificial Neural Networks (ANN) learning [8] [9] [10] [11]. Finding the best number of hidden neurons, learning rate, momentum coefficient and initial weights have been studied in the literature. Bovis *et al.* [12] worked on mammographic mass to find an optimum number of hidden neurons for classification. Mirjalili *et al.* [13] proposed a hybrid of PSO and gravitational search algorithm to train feed forward neural networks. PSO has been used to optimize the parameters of SVM. Bamakan *et al.* [14] proposed a hybrid approach for parameter determination of the non-parallel SVM using PSO. They considered the number of support vectors along with the classification accuracy as a weighted objective function.

In this study, we use PSO to optimize the number of hidden layers (L) and the number of neurons ($N_i$'s) in each layer for deep learning models [3]. To the best of our knowledge, no one has used PSO for setting these parameters. Currently, the $H_2O$ platform utilizes grid search for parameter selection. In our experiments, we observed that PSO results in a significant decrease in the number of configurations that need to be evaluated to find optimal parameters for deep learning models. Specifically the decrease was by 77% - 85% while achieving higher model accuracy compared to grid search. While the results presented in this paper are based on a dataset collected from a campus Wi-Fi network, we believe that PSO would result in similar results in other application domains.

The remainder of the paper is organized as follows: Section II presents the motivations behind this work. In Section III, we present our proposed deep learning parameter selection method using PSO. Section IV presents our experimental results and the lessons learned, and finally, Section V concludes this study and discusses future research directions.

## II. MOTIVATIONS

To the best of our knowledge, there is no theory yet to determine the best number of hidden layers and the number of neurons in each layer that should be used by a deep learning model to approximate a given function. There are several alternatives, rules of thumb that could mitigate the modelers' effort and time. For instance, number of hidden layers could be selected to be between the number of inputs and outputs [15]. Another rule suggests that the number of hidden layers can be based on the following formula [16]:

$$H \approx (I+O) \cdot 2/3 \qquad (1)$$

Where *H* is the number of neurons in the hidden layers, *I* is the number of features in the input layer, and *O* is the number of neurons in the output layer.

In [17], Swingler argues that the number of hidden layers should never exceed the number of input variables. In terms of neurons, the number of hidden layer neurons should be less than twice of the number of neurons in the input layer [18].

Configuring deep learning models using the above rules is almost free of any computations, since what all needed is a basic and simple calculation. However, these rules of thumb are not applicable all the time because they ignore the number of trainings, the amount of noise in the targets, and the complexity of the function. Further experiments using a large number of different datasets are needed in order to find good rules of thumb for the different application domains.

In our experiments, we use deep learning models for predictive modeling. $H_2O$ uses a purely supervised training protocol [19]. The configurations of deep learning algorithms in the $H_2O$ platform and other popular platforms have no default settings for the hidden layer size and the number of neurons in each layer. Experimenting with building deep learning models using different network topologies and different datasets will lead to intuition for these parameters. For manual parameter selection, we selected different configurations in our experiments in terms of the number of hidden layers and the number of neurons in each layer of the deep learning model. Figure 1 shows the effect of different configurations on the accuracy. The figure illustrates that the parameter selection process has a significant impact on the accuracy of the deep learning model. However, the number of potential configurations is large. In fact searching for the best configuration is like searching for a needle on a haystack.

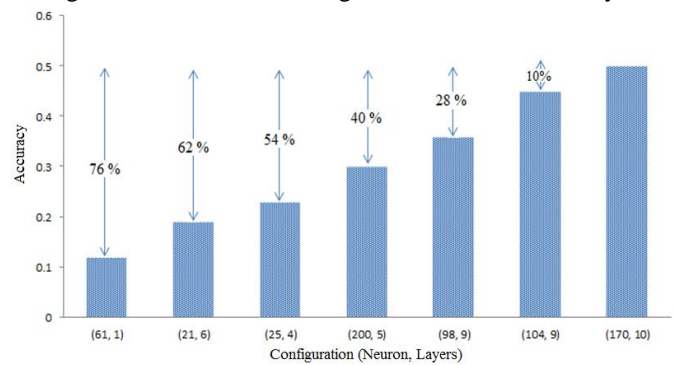

Fig. 1: The effect of manual configuration setting on the accuracy.

As illustrated in Figure 1, the number of hidden layers and the number of neurons in each layer play a major role to efficiently enhance the accuracy. For example, in comparison to the deep learning model that employs 10 hidden layers and 170 neurons in each layer, the accuracy is improved by 40 % for the deep learning model that employs 5 layers and 200 neurons per layer. This accuracy is further improved by 76 % when the deep learning model employs 1 hidden layer and 61 neurons in the layer. By running numerous configurations, one can find the best parameter values. However, that is a computationally intensive endeavor. Thus, it can be easily seen that finding high-quality parameter settings of a deep learning model is a time consuming process that requires an in-depth knowledge of the underlying algorithms, properties of the learning domain and the nature of the dataset that are being used in the training process. In section 4 of this paper, we compare our proposed PSO based parameter selection method with the grid search technique.

### III. PSO-BASED PARAMETER OPTIMIZATION MODEL

The PSO algorithm is an iterative optimization method that was originally proposed in 1995 by Kennedy and Eberhart [5]. PSO was developed to mimic bird and fish swarms. Animals who move as a swarm can reach their aims easily. The basic form of the PSO algorithm is composed of a group of particles which repeatedly communicate with each other; the population is called a swarm. Each particle represents a possible solution to the problem (i.e., the position of one particle represents the values of the attributes of a solution) [20]. Each particle has its position, velocity and a fitness value that is determined by an optimization function. The velocity determines the next direction and distance to move. The fitness value is an assessment of the quality of the particle. The position of each particle in the swarm is tweaked to move closer to the particle which has the best position. Each particle updates its velocity and position by tracking two extremes in each iteration. One is called the personal best, which is the best solution that the particle was able to obtain individually so far. The other is called the global best which is the best solution that all particles were able to find collectively so far.

PSO is mathematically modeled as follows [5]:

$$v_i^{t+1} = w \cdot v_i^t + c_1 \cdot rand \cdot (pbest_i - x_i^t) + c_2 \cdot rand \cdot (gbest - x_i^t) \quad (2)$$

Each step $t$, the position of particle $i$, $x_i^t$ is updated based on the particle's velocity $v_i^t$:

$$x_i^{t+1} = x_i^t + v_i^{t+1} \quad (3)$$

In Equations (2) and (3) above, $v_i^t$ and $x_i^t$ are the $t^{th}$ speed and position components of the $i^{th}$ particle. $c_1$ and $c_2$ are the acceleration coefficients and represent the weights of approaching the $pbest_i$ and gbest of a particle. $w$ is the inertia coefficient as it helps the particles to move by interia towards better positions. $rand$ is a uniform random value between 0 and 1. The parameters utilized in our experiments are listed in Table I.

TABLE I: THE PARAMETERS UTILIZED IN OUR EXPERIMENTS

| Parameter | Value |
|---|---|
| Population size | 10, 25, or 50 |
| Learning coefficients: c1, c2 | uniformly distributed between [0, 4] |
| Maximum number of iterations | 10 |
| Number of hidden layers | within the range [1, 200] |
| Number of neurons in each layer | within the range [1, 10] |
| Particle dimensions | represents the number of hidden layers and the number of neurons in each layer |
| Hidden layers velocity | MinLayerVelocity= -0.1(MaxLayers - MinLayers)<br>MaxNeuronVelocity= +0.1(MaxNeurons - MinNeurons) |
| Neuron velocity | MaxNeuronVelocity= 0.1 (MaxNeurons - MinNeurons)<br>MinNeuronVelocity= -(0.1 (MaxNeurons - MinNeurons)) |

---

**Algorithm 1**: PSO for Parameter Optimization of Deep Learning Models.

---

**Input:** Wi-Fi dataset, location, time and MAC addresses
**Output:** Optimal configuration in terms of the number of hidden layers and number of neurons in each layer for the deep learning model.
Begin:
1) Initialization
   a. Set the values of acceleration constants ($c_1$ and $c_2$), W, PopSize, MaxIt, and specify the range bounds: MinLayer, MaxLayer, MinNeurons, MaxNeurons, MaxLayerVelocity and MaxNeuronVelocity.
   b. Define the fitness function (i.e., deep learning model accuracy).
   c. Establish initial random population for the number of hidden layers and number of neurons in each layer.
   d. Calculate the fitness value for each particle and set the personal best (pbest) for each particle and the global best (gbest) for the population.
2) Repeat the following steps until the gbest solution does not change anymore or the maximum number of iterations is reached.
   a. Update the number of hidden layers, the number of neurons in each layer, the velocity of the number of hidden layers and the number of neurons in each particle according to the Equations (4) through (7).
   b. Calculate the fitness value for each particle. If the fitness value of the new location is better than the fitness value of personal best, the new location is updated to be the personal best location.
   c. If the currently best particle in the population is better than the global best, the best particle replaces the recorded global best.
3) Return the optimal number of hidden layers, the number of neurons in each layer for the deep learning model.
End

---

Algorithm 1 above provides the details of our proposed PSO based parameter selection techniques for deep learning models. The algorithm is presented for campus occupant prediction scenario using Wi-Fi collected data. This scenario will be fully explored in the next section (i.e., Section IV).

In our implementation of PSO, the $i^{th}$ particle's velocity is calculated according to the following:

- Velocity of number of layers

$$V_{L,i}^{t+1} = w \cdot V_{L,i}^t + c_1 \cdot rand \cdot (L_i^{best} - V_{L,i}^t) + c_2 \cdot rand \cdot (G^{Lbest} - V_{L,i}^t) \quad (4)$$

Where $V_L$ is the velocity of the number of hidden layers, $L_i^{best}$ is the particle's best local value of the number of hidden layers, and $G^{Lbest}$ is the best global value of the number of hidden layers.

- Velocity of number of neurons

$$V_{N,i}^{t+1} = w \cdot V_{N,i}^t + c_1 \cdot rand \cdot (N_i^{best} - V_{N,i}^t) + c_2 \cdot rand \cdot (G^{Nbest} - V_{N,i}^t) \quad (5)$$

Where $V_N$ is the velocity of the number of neurons in each hidden layer, $N_i^{best}$ is the particle's best local value of the number of neurons in each hidden layer, and $G^{Nbest}$ is the best global value of the number of neurons in each hidden layer.

- Position for number of layers
$$L_i^{t+1} = L_i^t + V_{L,i}^{t+1} \qquad (6)$$
- Position for number of neurons
$$N_i^{t+1} = N_i^t + V_{N,i}^{t+1} \qquad (7)$$

IV. EXPERIMENTAL RESULTS AND LESSONS LEARNED

In our experiments, we select a smart building application to assess the performance of our proposed PSO based parameter selection technique. We built a deep learning model based on 6 weeks (January 15, 2016 – Feb 29, 2016) of Wi-Fi access data collected from 14 buildings of the campus of the University of Houston campus. Our goal is to build a deep learning model that predicts the number of occupants at a given location in 15, 30 and 60 minutes from the current time. Awareness of the number of occupants in a building at a given time is crucial for many smart building applications including energy efficiency and emergency response services [21].

Our experiments were conducted using the R language. We executed our experiments on a 24-core machine with 2.40GHz Intel® Xeon® CPU and 32 GB RAM. In our scenarios, we split a 6 weeks dataset into 7 parts; each part corresponds to a day of the week. Each dataset has the following features: Access Point ID (APID), Date, Time, User MAC address and Building number. The three features that our deep learning model needs to predict are the count of MAC addresses within 15, 30 and 60 minutes from the current time at a given date, time and location (i.e., APID and Building number). In the process, we built a deep learning model for each day of the week. Table II summarizes the different parts of the dataset. Further, each dataset for a specific day of the week has been split into training and testing sets. Specifically, the first five weeks of the dataset were used as a training set while the data that pertains to the sixth week is used as a testing set. We then set out to address the main goal of this paper which is to compare our proposed PSO based parameter selection technique vis-à-vis the grid search technique in terms of finding the best parameters for the seven models that correspond to the days of the week.

In order to evaluate and compare the grid search and PSO approaches, both the accuracy and the number of configurations that need to be explored to get the best accuracy are evaluated. In case of PSO, the algorithm terminates when the maximum number of iterations is reached or when there is no difference between the accuracies of two consecutive iterations. Since the count of occupants at a given time and location is a continually changing number (i.e., regression problem), it does not make much sense to predict the exact number of occupants N at a given date, time and locations. Rather, it is more practical to allow a small tolerance in the count, for example N± n. Therefore, we consider clusters with window size ±n (e.g., 20) when we evaluate the accuracy of the predicted occupancy for each dataset.

Three different swarm sizes of 10, 25 and 50 particles are used in our experiments. Figure 2 shows the accuracy (c.f. Figure 2a) and the number of configurations that need to be evaluated (c.f. Figure 2b) to achieve that accuracy for predicting the occupancy within a 60 minute time window. These two figures jointly illustrate the by using a small population size (e.g., 10 particles), the PSO based parameter value selection technique was able to achieve an accuracy that is almost the same as that achieved by using a larger population size (e.g., 25 and 50 particles) for almost all the datasets that we experimented with. Therefore, the PSO based parameter value selection approach does not require a large number of particles to produce competitive results. Another observation drawn from Figure 2(b) is that the number of iterations needed to reach the globally best solution is almost one-third and one-fifth the number of configurations that need to be evaluated by the grid search method when the PSO based techniques employs 25 and 50 particles, respectively. This demonstrates that the PSO based technique can be computationally efficient to determine the deep learning parameters. Therefore, in the following experiments can simply consider the PSO based technique with 10 particles and compare our results with the grid searching technique.

Figures 3-5 show the number of different configurations that need to be evaluated to reach the globally best solution in terms of predicting the occupancy in the next 60, 30 and 15 minutes, respectively. These figures illustrate that better accuracy can be achieved when using our proposed PSO based parameter value selection technique while having to evaluate a significantly lower number of configurations compared to the grid search approach. This clearly exhibits the supremacy of the PSO based technique over grid search. Thus, it can serve as a great candidate for parameter tuning of deep machine learning models. Of course, one needs to carefully analyze dataset biases or domain specific properties that give rise to these results, but that is beyond the scope of this paper and is left for future extensions.

TABLE II: TRAINING AND TESTING DATASETS FOR THE DAYS OF THE WEEK

| Dataset | Number of records in training set | Number of records in testing set | Actual occupancy in the next 60 minute | Actual occupancy in the next 30 minute | Actual occupancy in the next 15 minute |
|---|---|---|---|---|---|
| Sat. | 335137 | 71551 | 167 | 110 | 93 |
| Sun. | 213434 | 108597 | 184 | 100 | 80 |
| Mon. | 1686200 | 795439 | 715 | 648 | 488 |
| Tue. | 2129033 | 411025 | 732 | 628 | 474 |
| Wed. | 2141754 | 404023 | 792 | 618 | 481 |
| Thur. | 1986703 | 269976 | 794 | 689 | 493 |
| Fri. | 1200046 | 253995 | 323 | 262 | 234 |

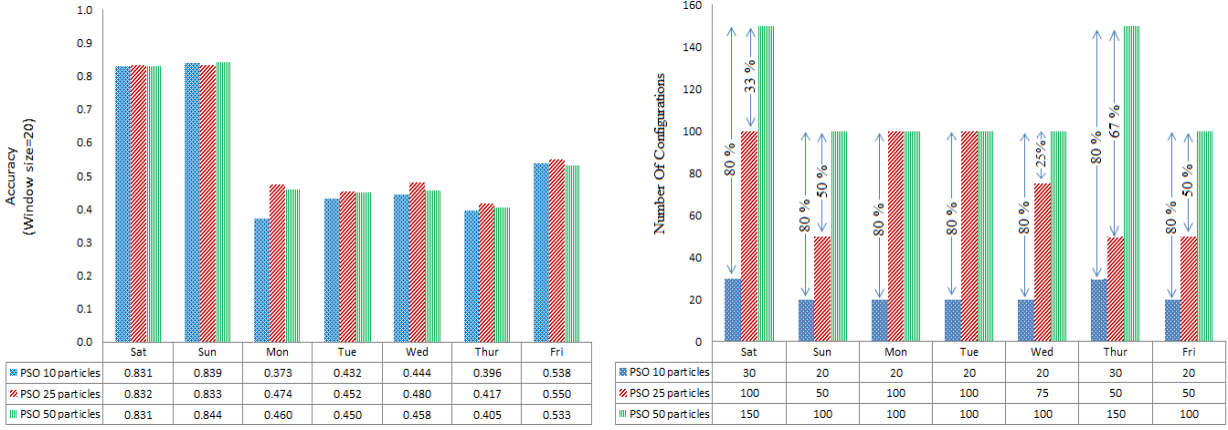

(a) Accuracy vs. Model  (b) Number of different configurations vs. Model
Fig. 2: Comparison between three different swarm sizes (10, 25 and 50).

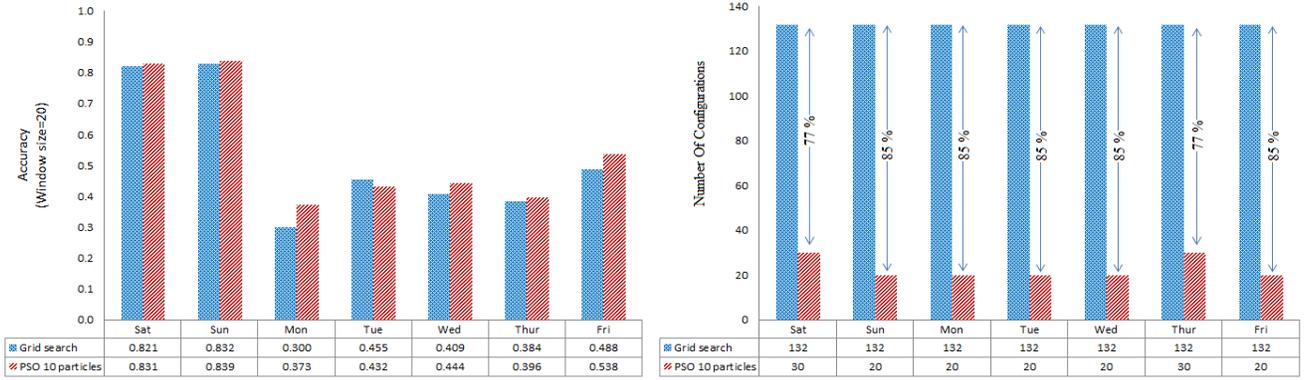

(a) Accuracy vs. Model  (b) Number of different configurations vs. Model
Fig. 3: Comparison between PSO and grid search to predict within a 60 minute interval.

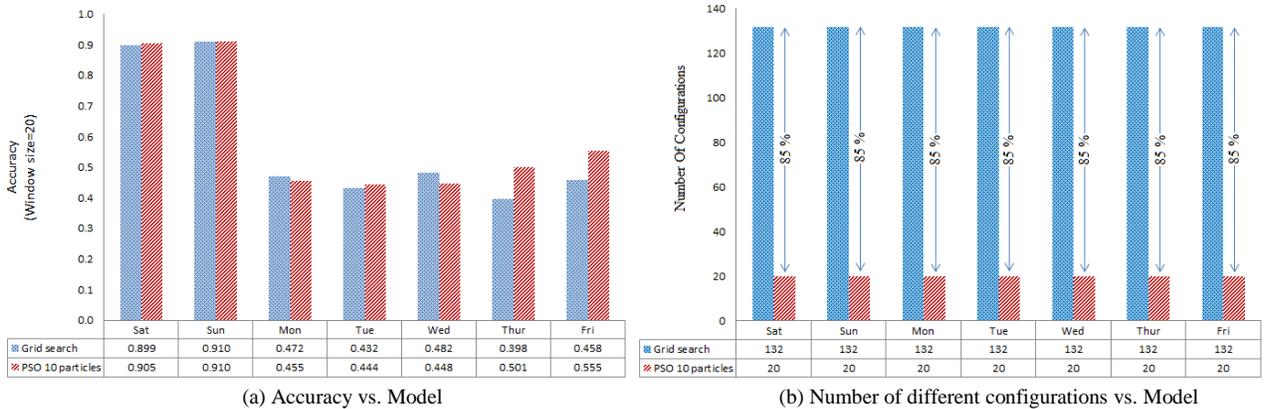

(a) Accuracy vs. Model  (b) Number of different configurations vs. Model
Fig. 4: Comparison between PSO and grid search to predict within a 30 minute interval.

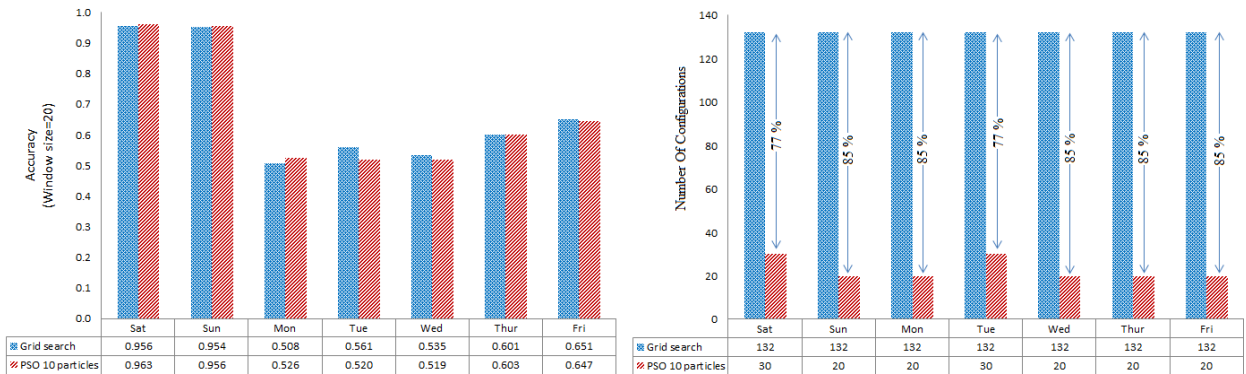

(a) Accuracy vs. Model  (b) Number of different configurations vs. Model
Fig. 5: Comparison between PSO and grid search to predict within a 15 minute interval.

## V. Conclusions and future work

Multiple parameters have to be set and tuned for deep learning models. These parameters can have a significant influence on the results and the computational needs of deep learning models. Optimization methods therefore need to be used to help find optimal parameter settings. Consequently, the user can focus on the results of deep learning rather than on spending time and efforts on deciding the optimal parameter values. This paper presents a PSO based parameter value selection technique to optimize the performance of deep learning models, by selecting the number of hidden layers and the number of neurons in each layer. Our results show that the proposed PSO algorithm is useful in the process of training deep learning models. We demonstrated the performance of the proposed technique in a smart building scenario where the number of occupants needs to be predicted in the next 60, 30 and 15 minutes based on collected Wi-Fi data. The results obtained show that training times decreased by 77% - 85% when using the PSO based approach compared to the grid search method. Our proposed PSO based technique also gives a better classification accuracy compared to the grid search approach. As a future extension, we intend to explore the use of PSO to tune other deep learning parameters such as: the activation functions and the number of epochs. Note that it is easy to implement parallel versions of PSO on GPUs. Therefore, resulting in further reduced training times, while letting researchers focus on extracting subject matter knowledge using deep learning models, rather than letting them focus on the parameter value selection process itself.

## Acknowledgment

This article was made possible by NPRP grant # [7-1113-1-199] from the Qatar National Research Fund (a member of Qatar Foundation). The statements made herein are solely the responsibility of the authors.

## References

[1] I. Goodfellow, Y. Bengio, and A. Courville, *Deep Learning*. Cambridge, MA: The MIT Press, 2016.
[2] "Machine Learning and Understanding for Intelligent Extreme Scale Scientific Computing and Discovery," Advanced Scientific Computing Research (ASCR) Division of the Office of Science, U.S. Department of Energy, Workshop Report, Jan. 2015. [Online]. Available: https://www.orau.gov/machinelearning2015/Machine_Learning_DOE_Workshop_Report_6.pdf . [Accessed: January 28-2017].
[3] Y. Malitsky, D. Mehta, B. O'Sullivan, and H. Simonis, "Tuning parameters of large neighborhood search for the machine reassignment problem," in *International Conference on AI and OR Techniques in Constraint Programming for Combinatorial Optimization Problems*, 2013, pp. 176–192.
[4] Y. Ganjisaffar, T. Debeauvais, S. Javanmardi, R. Caruana, and C. V. Lopes, "Distributed tuning of machine learning algorithms using MapReduce clusters," in *Proceedings of the Third Workshop on Large Scale Data Mining: Theory and Applications*, 2011, p. 2.
[5] J. Kennedy and R. Eberhart, "Particle swarm optimization," in *Proceedings of the IEEE international conference on neural networks IV*, 1995, pp. 1942-1948.
[6] C. W. de Silva, *Mechatronic Systems: Devices, Design, Control, Operation and Monitoring*. Boca Raton: CRC Press, 2007.
[7] J. Karwowski, M. Okulewicz, and J. Legierski, "Application of Particle Swarm Optimization Algorithm to Neural Network Training Process in the Localization of the Mobile Terminal," in *Engineering Applications of Neural Networks*, 2013, pp. 122–131.
[8] Y. M. M. Hassim and R. Ghazali, "Solving a classification task using Functional Link Neural Networks with modified Artificial Bee Colony," in *2013 Ninth International Conference on Natural Computation (ICNC)*, 2013, pp. 189–193.
[9] H. Shah and R. Ghazali, "Prediction of Earthquake Magnitude by an Improved ABC-MLP," in *2011 Developments in E-systems Engineering*, 2011, pp. 312–317.
[10] L. Qiongshuai and W. Shiqing, "A hybrid model of neural network and classification in wine," in *2011 3rd International Conference on Computer Research and Development*, 2011, vol. 3, pp. 58–61.
[11] B. A. Garro, H. Sossa and R. A. Vázquez, "Artificial neural network synthesis by means of artificial bee colony (ABC) algorithm," *2011 IEEE Congress of Evolutionary Computation (CEC)*, New Orleans, LA, 2011, pp. 331-338. doi: 10.1109/CEC.2011.5949637.
[12] K. Bovis, S. Singh, J. Fieldsend and C. Pinder, "Identification of masses in digital mammograms with MLP and RBF nets," in *Proceedings of the IEEE-INNS-ENNS International Joint Conference on Neural Networks. IJCNN 2000. Neural Computing: New Challenges and Perspectives for the New Millennium*, Como, 2000, pp. 342-347.
[13] S. Mirjalili, S. Z. Mohd Hashim, and H. Moradian Sardroudi, "Training feedforward neural networks using hybrid particle swarm optimization and gravitational search algorithm," *Applied Mathematics and Computation*, vol. 218, no. 22, Jul. 2012, pp. 11125–11137.
[14] S. M. H. Bamakan, H. Wang, and A. Z. Ravasan, "Parameters Optimization for Nonparallel Support Vector Machine by Particle Swarm Optimization," *Procedia Computer Science*, vol. 91, 2016, pp. 482–491.
[15] A. Blum, *Neural Networks in C++: An Object-Oriented Framework for Building Connectionist Systems*, 1 edition. New York: Wiley, 1992.
[16] Z. Boger and H. Guterman, "Knowledge extraction from artificial neural network models," in *Systems, Man, and Cybernetics, 1997. Computational Cybernetics and Simulation, 1997 IEEE International Conference on*, 1997, vol. 4, pp. 3030–3035.
[17] K. Swingler, Applying Neural Networks: A Practical Guide, Pap/Dsk edition. San Francisco: Morgan Kaufmann, 1996.
[18] G. S. Linoff and M. J. A. Berry, *Data Mining Techniques: For Marketing, Sales, and Customer Relationship Management*, 3 edition. Indianapolis, IN: Wiley, 2011.
[19] A. Arora, A. Candel, J. Lanford, E. LeDell, and V. Parmar, *Deep Learning with H$_2$O*. 2015.
[20] C. J. A. Bastos-Filho, D. F., M. P., P. B. C. Miranda, and E. M. N. Figueiredo, "Multi-Ring Particle Swarm Optimization," in *Evolutionary Computation*, W. P. dos Santos, Ed. InTech, 2009.
[21] V. L. Erickson, M. Á. Carreira-Perpiñán, and A. E. Cerpa, "Occupancy Modeling and Prediction for Building Energy Management," *ACM Trans. Sen. Netw.*, vol. 10, no. 3, May 2014, p. 42:1–42:28.